\pdfoutput=1

\documentclass[11pt]{article}

\usepackage{acl}
\usepackage{textgreek}

\usepackage{times}
\usepackage{latexsym}
\usepackage{xcolor}
\usepackage{color,soul}
\usepackage{booktabs}
\usepackage{multirow}
\usepackage{graphicx}
\usepackage{xytree}
\usepackage{tikz-dependency}
\usepackage{amsmath}
\usepackage{amssymb}
\usepackage{fdsymbol}
\usepackage{listings}
\usepackage[autostyle]{csquotes} 
\usepackage{mdframed}
\usepackage{subcaption}
\usepackage[normalem]{ulem}
\usepackage{scrextend}
\usepackage{fancyvrb}
\usepackage{enumitem}
\usepackage{tikz}
\usetikzlibrary{patterns}
\usepackage{pgfplots}
\pgfplotsset{compat=1.7}
\definecolor{colorone}{RGB}{90,180,172}
\definecolor{colortwo}{RGB}{199,234,229}
\definecolor{colorthree}{RGB}{1,102,94}

\usepackage[T1]{fontenc}

\usepackage[utf8]{inputenc}

\newcommand\blfootnote[1]{%
  \begingroup
  \renewcommand\thefootnote{}\footnote{#1}%
  \addtocounter{footnote}{-1}%
  \endgroup
}

\usepackage{microtype}

\title{A Dataset for N-ary Relation Extraction of Drug Combinations}

\author{Aryeh Tiktinsky$^\bigstar$^* \And
  Vijay Viswanathan$^\clubsuit$^* \AND
  Danna Niezni$^\spadesuit$ \And 
  Dana Meron Azagury$^\spadesuit$ \And
  Yosi Shamay$^\spadesuit$ \AND
  Hillel Taub-Tabib$^\bigstar$ \And 
  Tom Hope$^\bigstar$ \And 
  Yoav Goldberg$^\bigstar ^\varheartsuit$ \AND
  $^\bigstar$ {\normalfont Allen Institute for AI}, $^\clubsuit$ {\normalfont Language Technologies Institute, Carnegie Mellon University}, \\
  $^\spadesuit$  Faculty of Biomedical Engineering, Technion – Israel Institute of Technology, \AND
  \normalfont $^\varheartsuit$ Computer Science Department, Bar Ilan University\\
  }

 \author{
\textbf{Aryeh Tiktinsky}$^{*, 1}$
  \quad
\textbf{Vijay Viswanathan}$^{*, 2}$  \quad
\textbf{Danna Niezni}$^{3}$  \quad
\textbf{Dana Meron Azagury}$^{3}$ \\
\textbf{Yosi Shamay}$^{3}$ \quad
\textbf{Hillel Taub-Tabib}$^{1}$ \quad
\textbf{Tom Hope}$^{1, 4}$ \quad
\textbf{Yoav Goldberg}$^{1, 5}$ \quad
\\
$^1$Allen Institute for AI\quad
$^2$Language Technologies Institute, Carnegie Mellon University\quad\\
$^3$Faculty of Biomedical Engineering, Technion – Israel Institute of Technology\quad\\
$^4$School of Computer Science \& Engineering, The Hebrew University of Jerusalem\quad\\
$^5$Computer Science Department, Bar Ilan University\quad\\
{\small \texttt{aryeht@allenai.org} \phantom{aaaa}\texttt{vijayv@andrew.cmu.edu}} 
}

\def\bcbaux#1#2 #3\endbcb{%
  \colorbox{#1}{\strut#2}%
  \ifx\relax#3\relax\def\next{}\else%
    \colorbox{#1}{ \strut}%
    \allowbreak%
    \def\next{\bcbaux{#1}#3\endbcb}%
  \fi%
  \next%
}

\begin{document}
\maketitle
\VerbatimFootnotes
\begin{abstract}
\blfootnote{* Equal contribution.} 
Combination therapies have become the standard of care for diseases such as cancer, tuberculosis, malaria and HIV.  However, the combinatorial set of available multi-drug treatments creates a challenge in identifying effective combination therapies available in a situation.
To assist medical professionals in identifying beneficial drug-combinations, we construct an expert-annotated dataset for extracting information about the efficacy of drug combinations from the scientific literature. Beyond its practical utility, the dataset also presents a unique NLP challenge, as the first relation extraction dataset consisting of variable-length relations.  Furthermore, the relations in this dataset predominantly require language understanding beyond the sentence level, adding to the challenge of this task. We provide a promising baseline model and identify clear areas for further improvement. 
We release our dataset,\footnote{\url{https://huggingface.co/datasets/allenai/drug-combo-extraction}} code,\footnote{\url{https://github.com/allenai/drug-combo-extraction}} and baseline models\footnote{\url{https://huggingface.co/allenai/drug-combo-classifier-pubmedbert-dapt}} publicly to encourage the NLP community to participate in this task.
\end{abstract}

\begin{figure*}[!ht]
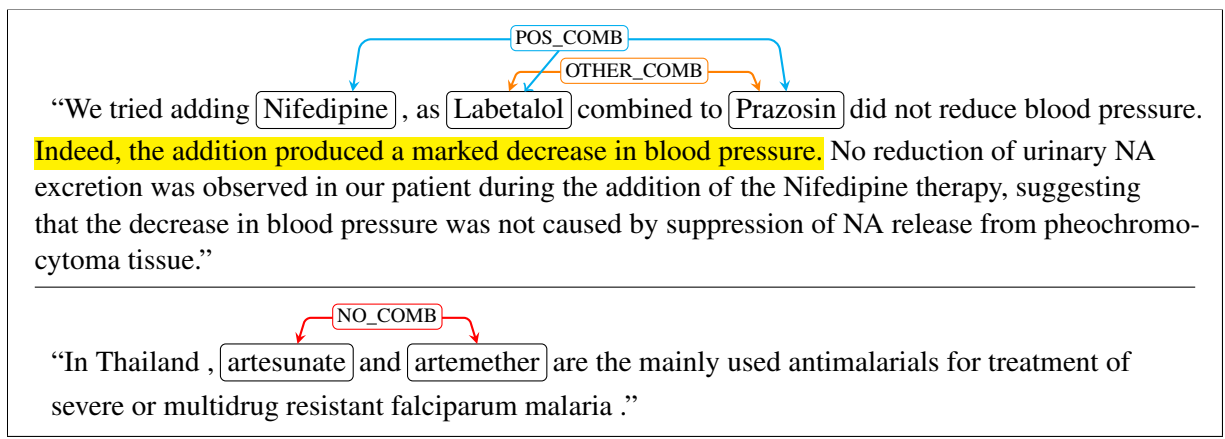


\begin{mdframed}
\begin{dependency}[theme = default, label style={draw=cyan}, edge style={thick, cyan}, edge horizontal padding=10pt]
   \begin{deptext}
      ``We tried adding \& Nifedipine \& , as \& Labetalol \& combined to \& Prazosin \& did not reduce blood pressure. \\
   \end{deptext}
    \depedge[label style={draw=orange}, edge height=1.5ex, edge style={orange, solid, <->}]{6}{4}{OTHER\_COMB}
    \depedge[label style={draw=cyan}, edge height=4ex, edge style={cyan, solid, <->}]{2}{6}{POS\_COMB}
    \storelabelnode\firstlab
    \draw [->, thick, cyan] (\firstlab)--(\wordref{1}{4});
    \wordgroup{1}{2}{2}{a0}
    \wordgroup{1}{4}{4}{a1}
    \wordgroup{1}{6}{6}{a2}
\end{dependency}\vspace{-5pt}
    \hl{Indeed, the addition produced a marked decrease in blood pressure.} No reduction of urinary NA \\ 
    excretion was observed in our patient during the addition of the Nifedipine therapy, suggesting \\
    that the decrease in blood pressure was not caused by suppression of NA release from pheochromocytoma tissue.''\\
\vspace{-20pt}

\hrulefill

\vspace{5pt}
\begin{dependency}[theme = default, label style={draw=cyan}, edge style={thick, cyan}]
   \begin{deptext}
      ``In Thailand , \& artesunate \& and \& artemether \& are the mainly used antimalarials for treatment of \\
   \end{deptext}
       \depedge[label style={draw=red}, edge height=2ex, edge style={red, solid, <->}]{2}{4}{NO\_COMB}
    \wordgroup{1}{2}{2}{a0}
    \wordgroup{1}{4}{4}{a1}
\end{dependency}
\vspace{-5pt}

\begin{dependency}[theme = default, label style={draw=cyan}, edge style={thick, cyan}]
   \begin{deptext}
      severe or multidrug resistant falciparum malaria .'' \\
   \end{deptext}
\end{dependency}
\vspace{-5pt}

\end{mdframed}
\caption{Examples of our label scheme. The top example contains two relations: a binary OTHER\_COMB relation and a ternary POS\_COMB relation. The evidence required to annotate the latter relation is found in a different sentence (highlighted). In the bottom example, each drug is described as a separate treatment rather than a combination therapy.}
\label{fig:main}
\end{figure*}

\section{Introduction}

``\textit{So far, many monotherapies have been tested, but have been shown to have limited efficacy against \textbf{COVID-19}. By contrast,
\textbf{combinational} therapies are emerging as a useful tool to treat
SARS-CoV-2 infection.}'' \citep{v13091768}.

Indeed, combining two or more drugs together 
has proven to be useful for treatments of various medical conditions, including
cancer \citep{PMID:162854,carew2008histone,shuhendler2010novel}, AIDS \citep{bartlett2006updated}, malaria \citep{eastman2009artemisinin}, tuberculosis \citep{bhusal2005determination}, hypertension \citep{rochlani2017two} and COVID-19 \citep{ianevski2020identification}.

In this work, we examine the clinically significant and challenging NLP task of extracting known
drug combinations from the scientific literature. We present an expert-annotated dataset and
 baseline models for this new task. 
Our dataset contains 1600 manually annotated abstracts, each mentioning between 2 and 15 drugs. 840 of these abstracts describe one or more positive drug combinations, varying in size from 2 to 11 drugs. The remaining 760 abstracts either contain mentions of drugs not used in combination, or discuss combinations of drugs that do not give a combined positive effect.

For the clinical setting, solving the drug combination
identification task can help researchers suggest and validate complex treatment plans. For example, when searching for effective treatments for cancer, knowing which drugs interact synergistically with a first line treatment allows researchers to suggest new treatment plans that can subsequently be validated in-vivo and become a standard protocol \citep{wasserman2001irinotecan,katzir2019prediction,ianevski2020identification,niezni:2022}.

From an NLP perspective, the drug combination identification task and dataset pushes the boundaries of relation
extraction (RE) research, by introducing a relation extraction task with several challenging
characteristics:
\\\noindent\textbf{Variable-length n-ary relations} Most work on relation
extraction is centered on \emph{binary relations} (e.g. \citet{cdr}, see full listing in \S\ref{related-work}), or on \emph{n-ary
relations with a fixed} $n$ (e.g. \citet{peng2017crosssentence}). In contrast, the drug combination task
involves \emph{variable-length n-ary relations}: different passages discuss
combinations of different numbers of drugs. For each
subset of drugs mentioned in a passage, the model must predict if they are used together in a combination therapy and whether this drug combination is effective.
\\\noindent\textbf{No type hints} As noted by \citet{rosenman2020exposing} and \citet{sabo2021revisiting}, in many relation extraction benchmarks \citep{han2018fewrel,sabo2021revisiting,zhang-etal-2017-position}, the argument types serve as an
effective clue. However, argument types do not apply naturally to the drug combination
task, in which all possible relation arguments are entities of the same type (drugs) and we
need to identify specific subsets of them.
\\\noindent\textbf{Long range dependencies} The information describing the efficacy of a
combination is often spread-out across multiple sentences. Indeed, our
annotators reported that for 67\% of the instances, the label could not be determined
based on a single sentence, requiring reasoning with a larger textual context.
Interestingly, our experiments show that our models \emph{are not} helped by
the availability of longer context, showing the limitations of current standard
modeling approaches. This suggests our dataset can be a test-bed for models that
attempt to incorporate longer context.
\\\noindent\textbf{Challenging inferences} As we show in our qualitative analysis
(\S\ref{section:error_analysis}), instances in this dataset require processing a range of
phenomena, including coordination, numerical reasoning, and world knowledge.

We hope that by releasing this dataset we will encourage NLP researchers to
engage in this important clinical task, while also pushing the boundaries of
relation extraction.

\section{The Drug Combinations Dataset}

A set of drugs in a biomedical abstract are classified to one of the following labels:
\\[0.5em]\noindent\textbf{Positive combination (POS\_COMB):} the sentence indicates the drugs are used in combination, and the passage suggests that the combination has additive, synergistic, or otherwise beneficial effects which warrant further study.
\\[0.5em]\noindent\textbf{Non-positive combination (OTHER\_COMB):} the sentence indicates the drugs are used in combination, but there is no evidence in the passage that the effect is positive (it is either negative or undetermined).\footnote{\label{neg-foot}We also experimented with another label for combinations that are discouraged (antagonistic, harmful or not effective). The agreement for this label was low, leading us to keep it as a subset of OTHER\_COMB.}
\\[0.5em]\noindent\textbf{Not a combination (NO\_COMB):}  the sentence does not state that the given drugs are used in combination, even if a combination is indicated somewhere else in the wider context. An example is given in the lower half of \autoref{fig:main}, where each of the drugs Artesunate and Artemether is given in isolation, and no combination is reported.

Our primary interest is to identify sets of drugs that are positive combinations.

\subsection{Relevant Context Size for Classifying Drug Combinations}
\label{ref:relevant_context_size}
When formulating the extraction task and designing our data collection methodology, we first analyzed the locality of the phenomenon: to what extent are drug combinations are expressed in a single sentence, or is a larger context is needed? We sampled $275$ abstracts that contained known drug combinations according to DrugComboDB.\footnote{We used \verb|Syner&Antag_voting.csv| taken from \url{http://drugcombdb.denglab.org/download/} and ranked according to the \textit{Voting} metric.} Analysis showed that 51\% of these abstracts mentioned attempted drug combinations. In 97\% of the abstracts containing drug combinations, all participating drugs in the attempted combination could be located within a single sentence in the abstract (for an example, see the OTHER\_COMB relation in \autoref{fig:main}).
However, establishing the efficacy of the combination frequently required a larger context (such as the context accompanying the POS\_COMB relation in \autoref{fig:main}). 

\subsection{Task Definition}

We define each instance in the Drug Combination Extraction (DCE) task to consist of a sentence, drug mentions within the sentence, and an enclosing context (e.g. paragraph or abstract). 


The output of the task is a set of relations, each consisting of a set of participating drug spans and a relation label (POS\_COMB or OTHER\_COMB). Each subset of drug mentions not included in the output set is implicitly considered to have relation label NO\_COMB.

More formally, DCE is the task of labeling an instance $X=\{C, i, D\}$ with a set of relation instances $R$, where $C=(S_1, ... S_n)$ is an ordered list of context sentences (e.g. all the sentences in an abstract or paragraph), $1 \leq i \leq n$ is an index of a target sentence $S_i=(w_1, ..., w_{n(i)})$ with $n(i)$ words, and $D=\{({d_\textbf{1}}_{start}, {d_\textbf{1}}_{end}), ..., ({d_\textbf{m}}_{start}, {d_\textbf{m}}_{end})\}$ is a set of $m>=2$ spans of drug mentions in $S$. The output is a set $R=\{(c_i, y_i)\}$ where $c_i \in \mathcal{P}(D)$ is a drug combination from $\mathcal{P}(D)$, the set of all possible drug combinations, and $y_i \in \{\text{POS\_COMB}, \text{OTHER\_COMB}\}$ is a combination label.

\subsection{Evaluation Metric}
\label{sec:eval_metrics}
We consider two settings: ``Exact Match'', a strict version which considers identifying exact drug combinations, and ``Partial Match'', a more relaxed version which assigns partial credits to correctly identified subsets.

We use standard precision, recall and F1 metrics for both settings. For the partial-match case, we replace the binary 0 or 1 score for a given combination with a refined score: $shared\_drugs/total\_drugs$.
If there are multiple partial matches with gold relations, we take the one with maximum overlap.
We compute \textbf{recall} as $identified\_relations/all\_gold\_relations$, and  \textbf{precision} as $correct\_relations/identified\_relations$. 

We consider two metrics, the averaged Positive Combination F1 score which compares POS\_COMB to the rest, and the averaged Any Combination F1 score which counts correct predictions for any combination label (POS or OTHER) as opposed to NO\_COMB. The latter is an easier task, but still valuable for identifying drug combinations irrespective of their efficacy.

\subsection{Collecting Data for Annotation}
\label{sec:collecting_data}
\begin{figure*}[!ht]
    \centering
    \resizebox{\textwidth}{!}{%
    \includegraphics{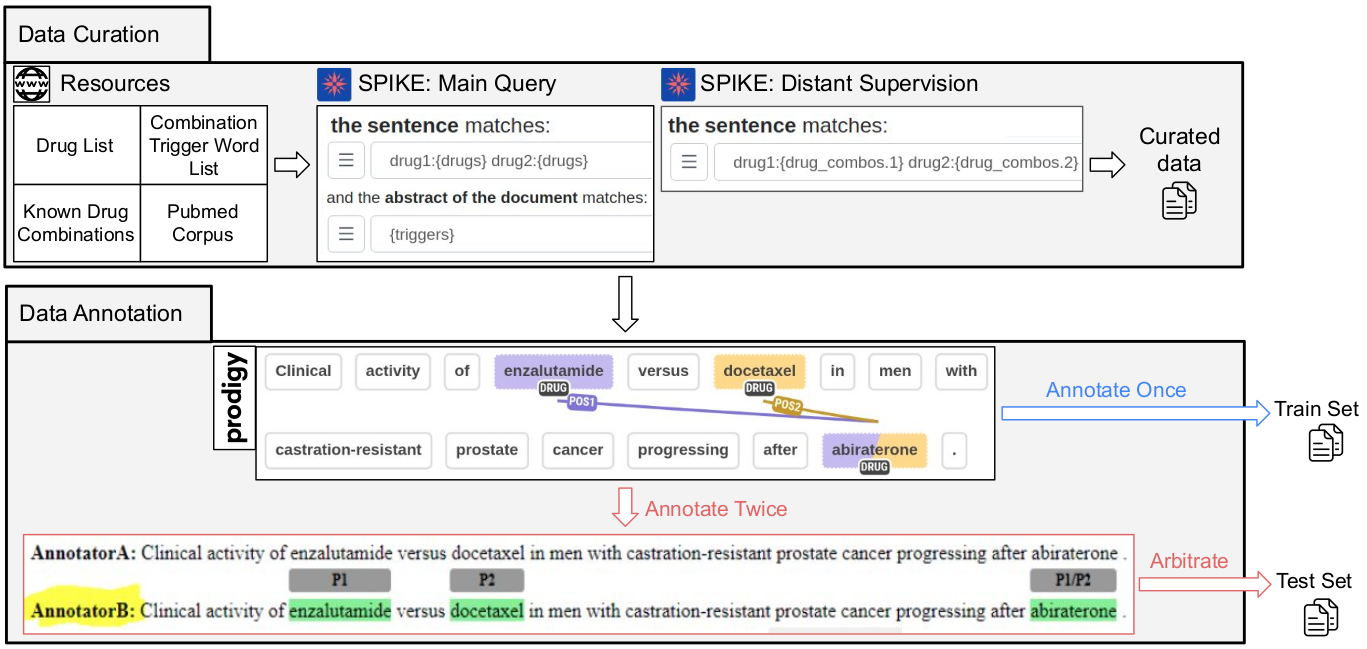}
    }
    \caption{Illustration of the data construction process. First we construct the required knowledge resources. Then, we collect data using SPIKE --an extractive search tool-- over the PubMed database. The train and test sets were annotated using Prodigy over the curated data. For test data, we collected two annotations for each sample, and then had a domain expert resolve annotation disagreements.}
    \label{fig:process}
\end{figure*}

To collect data for annotation we curated a list of 2411 drugs from DrugBank \footnote{Curation included downloading a premade drug list from DrugBank's website, while removing non pharmacological intervention such as Vitamins and Supplements. The later we got from the FDA orange book.} and sampled from PubMed a set of sentences which mention 2 or more drugs. Analysis of the first 50 sentences from this sample showed that only $8/50$ of the sentences included mentions of drug combinations. This meant that annotating the full sample will be costly, and will result in a dataset that's highly skewed toward relatively trivial NO\_COMB instances. 

We therefore repeated this experiment, sampling sentences whose PubMed abstract included a trigger phrase.\footnote{Triggers were selected by manually identifying words and phrases which frequently appear in abstracts mentioning drug combinations. These are phrases like “combination”, “followed by”, “prior to”, etc. (see full list in Appendix \ref{trigger-list}). The triggers are recall oriented, so while a presence of a trigger increases the chances that an abstract mentions a drug combination, it is definitely not clearly indicative. Importantly, since we’re dealing with a wide context, the presence of a trigger in an abstract which includes multiple drugs does not mean the trigger is related to the drugs.} 48\% of 50 sampled sentences included mentions of drug combinations. Evaluating the coverage of the trigger list against a new sample of abstracts with known drug combinations showed that 90\% of these new  abstracts included one of the trigger words. This suggests our trigger list is useful for fetching label-balanced data, without prohibitively restricting coverage and diversity. 

Accordingly, we collected the majority of instances for annotation, 90\%, using a basic search for sentences that contain at least two different drugs and whose abstract contains one of the trigger phrases. To overcome the lexical restrictions imposed by our trigger list, we sample the remaining 10\% of instances using distant supervision: fetching sentences containing pairs of drugs known to be synergistic according DrugComboDB, but whose abstract does not include one of our trigger phrases. All data collecting queries were performed using the SPIKE Extractive Search tool \cite{Shlain2020SyntacticSB,taub2020interactive}. The process is illustrated at the top of \autoref{fig:process}.

\subsection{The Annotation Process}

Seven graduate students in biomedical engineering took part in the annotation task. The students all completed a course in combination therapies for cancer and were supervised by a principled researcher with expertise in this field. 

We provided the participants with annotation guidelines which specified how the annotation process should be carried out (see Appendix \ref{guidelines}) and conducted an initial meeting where we reviewed the guidelines with the group and discussed some of the examples together. 

Each of the participants had access to a separate instance of the Prodigy annotation tool \citep{Prodigy:2018}, pre-loaded with the candidate annotation instances. Once a session starts, the instances (containing of a sentence with marked drug entities, and its context) appear in a sequential manner, with no time limit. For each instance we instructed the annotators to mark all subsets of drugs that participated in a combination, and for each subset to indicate its label (POS\_COMB or OTHER\_COMB). Moreover, we instructed them to indicate whether the context was needed in order to determine the positive efficacy of the relation.


Despite the considerable time required for expert annotation, we collected annotations for 1634 passages.
Among these, 272 were assigned to at least two annotators. After further arbitration by the lead researcher, these were used for the test set. The process is illustrated in the bottom part of \autoref{fig:process}.

\subsection{Inter-annotator Agreement}



During the course of the task we calculated inter-annotator agreement multiple times to identify cases of disagreement and provide feedback to annotators. Each time, a set of 25 instances were randomly selected and assigned to all annotators. Agreement was calculated based on a pairwise F1 measure (with some modifications as described in \S\ref{sec:eval_metrics}) and averaged over all pairs of annotators (see discussion of alternative metrics in Appendix \ref{metric}). 

\begin{table}[!t]
    \begin{center}
    \scalebox{0.75}{
    \begin{tabular}{c  c c} 
     \hline
     Metric & Partial Match & Exact Match \\ 
     \hline
     Avg. Any Combination F1 & 88.9 & 86.1 \\
     Avg. Positive Combination F1 & 83.4 & 79.6 \\
     \hline
    \end{tabular}}
    \end{center}
    \caption{Agreement scores using our adaptation of F1 score to allow for partial-match.}
    \label{table:agreement-score}
\end{table}

Final agreement numbers, in \autoref{table:agreement-score}, are satisfactory \citep{aroyo2013measuring,araki2018interoperable}.

\begin{figure*}[!ht]
    \centering
    \resizebox{\textwidth}{!}{%
    \includegraphics{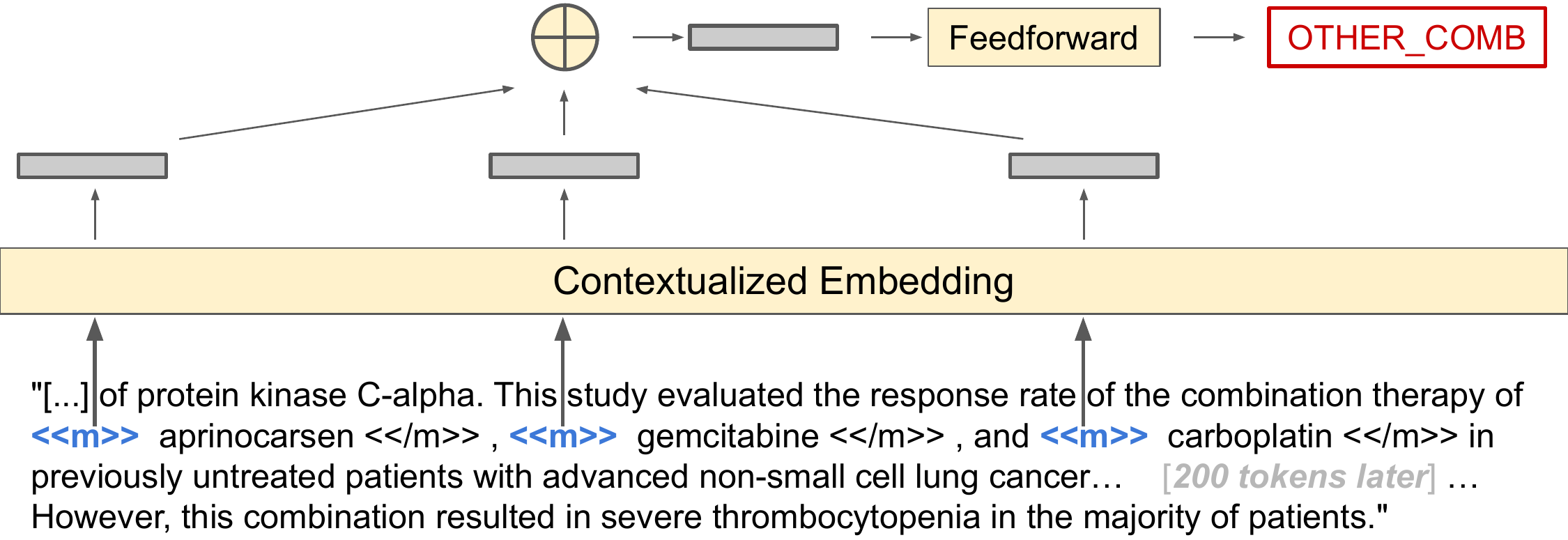}
    }
    \caption{Our baseline architecture, adapted from the PURE model \citep{zhong2021frustratingly}}
    \label{fig:baseline_model}
\end{figure*}

\subsection{Resulting Dataset}
The dataset consists of 1634 annotated abstracts,\footnote{This is a similar size to existing human-labeled biomedical relation extraction datasets, such as BioCreative V CDR \citep{cdr}, which has 1500 abstracts annotated, BioCreative VI \citep{Krallinger2017OverviewOT}, which has 2432 abstracts, and DDI \citep{herrero2013ddi}, which has 714 abstracts.} split into 1362 train and 272 test instances. These include 1248 relations; 838 are POS\_COMB and 410 are OTHER\_COMB (with the same label ratio in the train and test sets). 591 sentences contain no drug combination, 877 contain one relation (either POS\_COMB or OTHER\_COMB), and 166 contain two or more different combinations. Among annotated relations, 900 are binary, 226 are 3-ary, 69 are 4-ary, and 53 are 5-ary or more.

For each instance in the resulting dataset we include the context-required indication provided by the annotators. In 835 out of 1248 relations the annotator marked the context as needed which is 67\% of the time, showing the importance of the context in the DCE task.

\section{Experiments}

\begin{table*}[!ht] 
  \centering
  \small
  \begin{tabular}{lcccc} 
  \toprule
  \textbf{Model} & \multicolumn{2}{c}{Positive Combination F1} & \multicolumn{2}{c}{Any Combination F1} \\
   & Exact Match & Partial Match & Exact Match & Partial Match \\  \midrule

  Human-Level & 79.6  & 83.4  & 86.1   & 88.9  \\
  \midrule
  Rule-based & 31.8  & 45.6  & 39.1   & 57.4  \\
  \midrule
  SciBERT & 44.6 ($\pm$ 4.6) & 55.0 ($\pm$ 5.9)  & 50.2 ($\pm$ 1.9) & 63.6 ($\pm$ 2.7)  \\
  \hspace{10pt}w/ DAPT & 54.8 ($\pm$ 3.2) & 63.6 ($\pm$ 2.0) & 61.8 ($\pm$ 2.7) & 72.8 ($\pm$ 2.1)  \\

  BlueBERT & 41.2 ($\pm$ 4.8) & 51.7 ($\pm$ 6.0) & 47.3 ($\pm$ 4.2) & 59.9 ($\pm$ 6.2)  \\
  \hspace{10pt}w/ DAPT & 56.6 ($\pm$ 2.3) & 63.5 ($\pm$ 3.1) & 64.2 ($\pm$ 2.6) & 74.7 ($\pm$ 2.7)  \\
  PubmedBERT  & 50.7 ($\pm$ 5.5) & 59.6 ($\pm$ 5.8)  & 55.9 ($\pm$ 3.2) & 66.7 ($\pm$ 3.8)\\
  \hspace{10pt}w/ DAPT & \textbf{61.8} ($\pm$ 5.1) & \textbf{67.7} ($\pm$ 4.8) & \textbf{69.4} ($\pm$ 1.7) & \textbf{77.5} ($\pm$ 2.2)  \\
  BioBERT & 45.4 ($\pm$ 3.7) & 55.8 ($\pm$ 2.2)  & 46.7 ($\pm$ 3.6) & 58.3 ($\pm$ 5.1) \\ 
  \hspace{10pt}w/ DAPT & 56.0 ($\pm$ 6.5) & 63.5 ($\pm$ 7.5) & 65.6 ($\pm$ 1.8) & 75.7 ($\pm$ 2.2)  \\
  \bottomrule
  \end{tabular}
  \caption{Comparing different foundation models (with and without continued domain-adaptive pretraining) on Exact-Match and Partial-Match relation extraction metrics. Mean score from 4 different random seeds is reported, and standard deviation is computed across seeds.
  }
   \label{table:main-results} 
\end{table*}

\subsection{Baseline Model Architecture}

We establish a baseline model to measure the difficulty of our dataset and reveal areas for improvement. For our underlying baseline model architecture, we adopt the PURE architecture from \citet{zhong2021frustratingly}, which is state-of-the-art on several relation classification benchmarks, including the SciERC binary scientific RE dataset \citep{luan-etal-2018-multi}. The PURE architecture, designed for 2-ary and 3-ary relation extraction, consists of three components. First, special ``entity marker" tokens are inserted around all entities in a candidate relation. Next, these marker tokens are encoded with a contextualized embedding model. Finally, the entity marker embeddings are concatenated and fed to a feedforward layer for prediction.

Unlike the original PURE architecture, we consider the more challenging case of extracting relations of variable arity. To support this setting, we \emph{average} the entity marker tokens in a relation rather than concatenate. The final baseline model architecture is shown in \autoref{fig:baseline_model}. For the contextual embedding component of this architecture, we experiment with four different pretrained scientific language understanding models (SciBERT \citep{scibert}, BlueBERT \citep{bluebert}, PubmedBERT \citep{pubmedbert}, and BioBERT \citep{biobert}). During training, we only finetune the final *BERT layer. We train each model architecture for 10 epochs on a single NVIDIA Tesla T4 GPU with 15GB of GPU memory, which takes roughly 7 hours to train for each model.

To our knowledge, there are no other models designed for variable-length $N$-ary relation extraction, so we consider no other baselines.

\subsection{Domain-Adaptive Pretraining}
Our baseline model architecture relies heavily on a pretrained contextual embedding model to provide discriminative features to the relation classifier. \citet{gururangan-etal-2020-dont} showed that continued domain-adaptive pretraining almost always leads to significantly improved downstream task performance. Following this paradigm, we performed continued domain-adaptive pretraining (``DAPT'') on our contextual embedding models.

We acquired in-domain pretraining data using the same procedure used to collect data for annotation: running a SPIKE query against PubMed to find abstracts containing multiple drug names and a ``trigger phrase" (from the list in Appendix \ref{trigger-list}). This query resulted in 190K unique abstracts. We do not include any paragraphs from our annotated dataset. We then perform domain-adaptive training against this dataset using the \verb+Hugging Face Transformers+ library. We train for 10 epochs using a learning rate of 5e-4, finetuning all *BERT layers and using the same optimization parameters specified by \citet{gururangan-etal-2020-dont}. This pretraining took roughly 8 hours using four 15GB NVIDIA Tesla T4 GPUs.

\subsection{Relation Prediction}
To apply the model to drug combination extraction, we reduce the RE task to an RC task by considering all subsets of drug combinations in a sentence, treating each one as a separate classification input, and  combining the predictions.

This poses two challenges: there may be a large number of candidate relations for a given document, and each relation is classified independently despite the combinatorial structure. To handle these issues, we use a greedy heuristic of  choosing the smallest set of disjoint relations whose union covers as many drug entities as possible in the sentence. We do this iteratively: at each step, we choose the largest predicted relation that does not contain any drugs found in the relations chosen at previous iterations.

This greedy heuristic favors large (high arity) relations. Nonetheless, we empirically find this method is helpful for extracting high-precision drug combinations from our model architecture.

\subsection{Rule-based baseline}
To further validate that the trigger words do not introduce bias to our task, we consider an additional baseline based on the following rule: if a trigger word is found in the same sentence with multiple drugs, this set of drugs is tagged as POS\_COMB.

\section{Results}

\subsection{Effect of Pretrained LMs and Domain-Adaptive Pretraining}
We show results of our baseline model architectures in \autoref{table:main-results}. For each model, we report the mean and standard deviation of each metric over four identical models trained with different seeds.\footnote{Seeds used are 2021, 2022, 2023, and 2024} Among the four base scientific language understanding models in our experiments, we observe PubmedBERT to be the strongest on every metric. We additionally find that domain-adaptive pretraining provides significantly improvements for every base model, consistently giving 5-10 points of improvement on Positive Combination F1 score. The value of domain-adaptive pretraining supports our observation that encoding domain knowledge is critical to solving this new task. 

The rule-based approach underperformed all learned models (30 F1 points under our strongest model, PubmedBERT-DAPT). This shows this task cannot be reduced to keyword identification.


\subsection{Qualitative Error Analysis}

\begin{table*}[t] 
  \centering
  \small
  \begin{tabular}{lcccc}
  \toprule
  \textbf{Model} & \multicolumn{2}{c}{Positive Combination F1} & \multicolumn{2}{c}{Any Combination F1} \\
   & Exact Match & Partial Match & Exact Match & Partial Match \\  \midrule

  No Extra-sentential Context & 63.4 ($\pm$ 0.6) & 68.5 ($\pm$ 1.1) & 69.7 ($\pm$ 1.3) & 76.8 ($\pm$ 1.7)  \\
  1 Sentence of Context & 63.9 ($\pm$ 2.3) & 69.4 ($\pm$ 3.5)  & 71.9 ($\pm$ 1.1) & 78.6 ($\pm$ 1.8)  \\
  2 Sentences of Context  & 61.9 ($\pm$ 9.0) & 67.6 ($\pm$ 9.2)  & 70.1 ($\pm$ 2.3) & 77.9 ($\pm$ 1.3)\\
  3 Sentences of Context & 65.2 ($\pm$ 2.3) & 72.4 ($\pm$ 1.3)  & 70.8 ($\pm$ 1.7) & 78.7 ($\pm$ 1.2) \\ 
  \bottomrule
  \end{tabular}
  \caption{The effect of extra-sentential context on model performance. $n$ sentences are included on each side of the relation-bearing sentence. Mean and standard deviation of each metric are reported over 4 different random seeds.
  }
  \label{table:ctx_vs_not} 
\end{table*}

\label{section:error_analysis}
We identify classes of challenges that make this task difficult, both in terms of human annotation and machine prediction.
\\[0.5em]\noindent {\bf Coordination Ambiguity:} A known linguistic challenge is the ambiguity that stems from vague coordination. In cases where explicit combination words (e.g. combination, plus, together with, etc) are not used, it may be unclear whether two drugs are being used together or separately. For example in \textit{``These findings may help clinicians identify patients for whom acamprosate \textbf{and} naltrexone may be most beneficial''} it is unclear if \emph{acamprosate} and \emph{naltrexone} are being described in combination or as independent treatments, leading to either a POS label for the former or NO\_COMB for the latter.
\\[0.5em]\noindent{\bf Numerical and Relative Reasoning:} In some cases, the effect of a treatment is described in relative or numerical terms, rather than an absolute claim. Consider the example, \emph{``The infection rate in the control group was 3.5\% and in the treated group 0.5\%.''}. Here, the reader must compare the control vs experimental groups and deduce that the experimental outcome is positive, because the treatment yields a lower infection rate.
\\[0.5em]\noindent{\bf Domain Knowledge:} Similarly, classifying relations in this dataset may require an understanding of domain knowledge. In \textit{``Growth inhibition and apoptosis were significantly higher in BxPC-3, HPAC, and PANC-1 cells treated with \textbf{celecoxib and erlotinib} than cells treated with either \textbf{celecoxib} or \textbf{erlotinib}''}, one must understand that having higher values of \textit{Growth inhibition and apoptosis} in specific cells is a positive outcome, in order to classify this combination as positive.
\\[0.5em]\noindent{\bf Context related Complications:} The following are kinds of complications found when the evidence lies in the wider part of the context.
\\[0.5em]\noindent\underline{Coreference:} Anaphoric or coreferential reasoning is sometimes needed to understand the efficacy of the combination e.g. \textit{``it was demonstrated that \textbf{they} could be combined with acceptable toxicity.''}.
\\[0.5em]\noindent\underline{Contradicting Evidence:} the reader often must infer a conclusion given opposing claims within a given abstract. This can happen as combinations can be referred as e.g. \textit{toxic but effective}.
\\[0.5em]\noindent\underline{Long Distance:} The target sentence can be far---up to 41 sentences apart---from the evidence sentence, making it difficult for even humans to process.



\subsection{Quantitative Error Analysis}
To probe this task, we analyze the performance of our strongest model---the one using a PubmedBERT base model tuned with domain-adaptive pretraining---along different partitions of test data. We trained with four random seeds and perform comparisons using a paired multi-bootstrap hypothesis test where bootstrap samples are generated by sampling hierarchically over the four random seeds and the subsets of the test set \citep{Sellam2021TheMB}. We use 1000 bootstrap samples in each test.

\subsubsection{Do models leverage context effectively?}

Each relation in our dataset consists of entities contained within a single sentence, but labeling the relation frequently requires extra-sentential context to make a decision. In our dataset, annotators record whether or not each relation requires paragraph-level context to label, reporting that 67\% of drug combinations required context to annotate.

To understand the extent that models make use of paragraph-level context, we trained and evaluated our PubmedBERT-based model using varying amounts of extra-sentential context around the sentence containing drug entities. In \autoref{table:ctx_vs_not}, we see that adding context provides nearly identical performance to training a model with no extra-sentential context at all, with differences rarely exceeding one standard deviation of F1 score. 

%

However, we see increased variability in ``Positive Combination F1'' performance when extra-sentential context is used. To explain this, recall from \S\ref{ref:relevant_context_size} that determining the \emph{efficacy} of a drug combination often requires paragraph-level context for annotators, while identifying \emph{any combination} usually requires no context. From qualitative analysis of attention maps, we observed that our models are not able to consistently identify the salient parts of paragraph-level context, potentially causing instability with larger inputs.

These results suggest ample room for improvement in extracting document-level evidence. This makes our dataset a potentially useful benchmark for document-level language understanding. 

\subsubsection{Binary vs. higher-arity relations}
\label{sec:relation_arity}
Given that our dataset is the first relation extraction dataset with variable-arity relations, do higher-order relations pose a particular challenge for our models? To answer this, we separate all predicted and ground truth relations for the test set into binary relations and higher-arity relations. We then report precision among each subset of predicted relations and recall among each subset of ground truth relations.
\begin{figure}[t]
\centering
\resizebox{\linewidth}{!}{
    \begin{tikzpicture}
        \begin{axis}[
            ybar,
            ybar=7pt,
            ymin=0.4,
            ymax=1.0,
            enlarge x limits=0.2,
            x=80pt,
            bar width=24pt, 
            xlabel={Metric (for Positive Combination RE)},
            ylabel={Metric Value},
            symbolic x coords={Exact Precision, Exact Recall, Partial Precision, Partial Recall},
            xtick=data,
            nodes near coords,
            every node near coord/.append style={yshift=18pt,/pgf/number format/.cd, fixed, precision=2},
            legend cell align={left},
            xlabel near ticks,
            ylabel near ticks,
            ytick align=outside,
            ytick pos=left,
            xtick align=inside,
            xtick style={draw=none},
            ylabel shift=-5 pt,
            height=6.1cm,
            width=12.2cm,
            legend style={at={(0.02,0.97)},anchor=north west,nodes={scale=0.9, transform shape}}
            ]
            \addplot+ [
                colortwo!20!black,fill=colortwo!80!white,
                error bars/.cd,
                    y dir=both,
                    y explicit,
            ] coordinates {
                (Exact Precision,0.6016) +- (0,0.0465)
                (Exact Recall,0.6734) +- (0,0.0483)
                (Partial Precision,0.6459) +- (0,0.044)
                (Partial Recall,0.6996) +- (0,0.046)
            };

            \addplot+ [
                colorthree!20!black,fill=colorthree!80!white,
                error bars/.cd,
                    y dir=both,
                    y explicit, 
            ] coordinates {
                (Exact Precision,0.5897) +- (0,0.0783)
                (Exact Recall,0.5794) +- (0,0.0789)
                (Partial Precision,0.6995) +- (0,0.065)
                (Partial Recall,0.704) +- (0,0.0649)
            };
            \legend{Binary Relations, $N$-ary Relations ($N \geq 3)$}

        \end{axis}
        
    \end{tikzpicture}
    }
    \caption{Comparing models performance on binary vs higher-order $N$-ary relations, averaged over 4 seeds of the PubmedBERT-DAPT model. No consistent significant differences were observed; $p$-values for these comparisons are 0.456, 0.149, 0.240, and 0.276.}
    \label{fig:performance_by_arity}
    \vspace{-10px}
\end{figure}
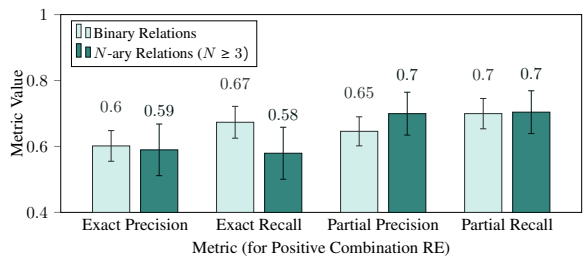
We perform this experiment across four different model seeds, and report results in aggregate using a paired multi-bootstrap procedure. In the results in \autoref{fig:performance_by_arity}, we see no consistent significant differences between models of different arities, suggesting that our technique of computing relation representations by averaging entity representations scales well to higher-order relations.

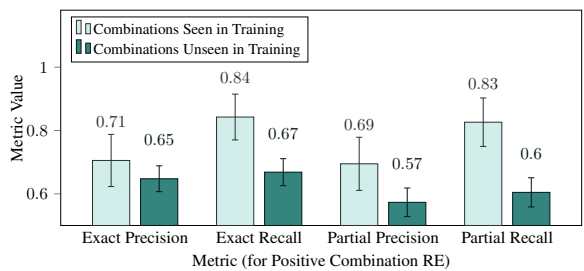
\begin{figure}[t]
\centering
\resizebox{\linewidth}{!}{
    \begin{tikzpicture}
        \begin{axis}[
            ybar,
            ybar=7pt,
            ymin=0.5,
            ymax=1.18,
            enlarge x limits=0.2,
            x=80pt,
            bar width=24pt, 
            xlabel={Metric (for Positive Combination RE)},
            ylabel={Metric Value},
            symbolic x coords={Exact Precision, Exact Recall, Partial Precision, Partial Recall},
            xtick=data,
            nodes near coords,
            every node near coord/.append style={yshift=18pt,/pgf/number format/.cd, fixed, precision=2},
            legend cell align={left},
            xlabel near ticks,
            ylabel near ticks,
            ytick align=outside,
            ytick pos=left,
            xtick align=inside,
            xtick style={draw=none},
            ylabel shift=-5 pt,
            height=6.5cm,
            width=12.2cm,
            legend style={at={(0.02,0.97)},anchor=north west,nodes={scale=0.9, transform shape}}
            ]
            \addplot+ [
                colortwo!20!black,fill=colortwo!80!white,
                error bars/.cd,
                    y dir=both,
                    y explicit,
            ] coordinates {
                (Exact Precision,0.7054) +- (0,0.0823)
                (Exact Recall,0.8425) +- (0,0.0723)
                (Partial Precision,0.6948) +- (0,0.0839)
                (Partial Recall,0.8262) +- (0,0.0765)
            };

            \addplot+ [
                colorthree!20!black,fill=colorthree!80!white,
                error bars/.cd,
                    y dir=both,
                    y explicit, 
            ] coordinates {
                (Exact Precision,0.6476) +- (0,0.0408)
                (Exact Recall,0.6684) +- (0,0.0426)
                (Partial Precision,0.5731) +- (0,0.0452)
                (Partial Recall,0.6046) +- (0,0.0461)
            };
            \legend{Combinations Seen in Training, Combinations Unseen in Training}.

        \end{axis}
        
    \end{tikzpicture}
    }
    \caption{Comparing relation extraction on test set drug combinations that are observed in the training set or not, using the PubmedBERT-DAPT model. Paired multi-bootstrap test $p$-values for these four comparisons are 0.262, 0.025, 0.103, and 0.009, respectively.}
    \label{fig:observed_in_training_set}
    \vspace{-10px}
\end{figure}

\subsubsection{Generalizing to new drug combinations}
\label{sec:unseen_generalization}
How well can relation extraction models classify drug combinations not seen during training? Similar to the setup in \S\ref{sec:relation_arity}, we divide all predicted and ground truth relations for the test set into the set of drug combinations which are also annotated in our training set, and the set that have not been. In our dataset, over 80\% of annotated test set relations are not found in the training set.

In \autoref{fig:observed_in_training_set}, performance is consistently better for relations observed in the training set than for unseen relations, by a margin of 10-15 points. Recall, in particular, is significantly worse for relations unseen during training (at 95\% confidence), and precision is potentially also worse. Considering that unseen drug combinations are practically more valuable than already-known combinations, improving generalization to new combinations is a critical area of improvement for this task.

\section{Related Work}
\label{related-work}

The DDI dataset \citep{herrero2013ddi} is the only work to our knowledge that annotates drug interactions for text mining. However, it fundamentally differs from our dataset in the type of annotations provided: the DDI annotates the type of discourse context in which a drug combination is mentioned, without providing explicit information about combination efficacy. In contrast, our dataset is focused on semantically classifying the efficacy of drug combinations as stated in text.

Other RE datasets exist in the biomedical field \citep{peng2017crosssentence,cdr,wu2019renet,Krallinger2017OverviewOT}, but do not focus on drug combinations. Similarly, several RE datasets tackle the $N$-arity problem in the scientific domain \citep{peng2017crosssentence,jain2020scirex,kardas-etal-2020-axcell,hou-etal-2019-identification}, and in the non-scientific domain \citep{akimoto-etal-2019-cross,nguyen-etal-2016-dataset}, however, \textbf{all of them consider a fixed choice of $N$}.

\section{Conclusions}

We present a new resource for drug combination and efficacy identification. We establish baseline models that achieve promising results but reveal clear areas for improvement. Beyond the immediate, application-ready value of this task, this task poses unique relation extraction challenges as the first dataset containing variable-arity relations. We also highlight challenges with document-level representation learning and incorporating domain knowledge. We encourage others to participate in this task. Our processed dataset\footnote{\url{https://huggingface.co/datasets/allenai/drug-combo-extraction}}
 and best baseline model\footnote{\url{https://huggingface.co/allenai/drug-combo-classifier-pubmedbert-dapt}} are available on Hugging Face, and our model training code
is available to the public at \url{https://gi~thub.com/allenai/drug-combo-extraction}.

\section*{Acknowledgements}

This project has received funding in part from the European Research Council (ERC) under the European Union's Horizon2020 research and innovation programme, grant agreement   802774 (iEXTRACT), and in part from the NSF Convergence Accelerator Award \#2132318.
We would also like to thank our annotators from the Shamay lab at the Faculty of Biomedical Engineering, Technion, including Shaked Launer-Wachs, Yuval Harris, Maytal Avrashami, Hagit Sason-Bauer and Yakir Amrusi

\bibliography{custom}

\begin{thebibliography}{42}
\expandafter\ifx\csname natexlab\endcsname\relax\def\natexlab#1{#1}\fi

\bibitem[{Akimoto et~al.(2019)Akimoto, Hiraoka, Sadamasa, and
  Niepert}]{akimoto-etal-2019-cross}
Kosuke Akimoto, Takuya Hiraoka, Kunihiko Sadamasa, and Mathias Niepert. 2019.
\newblock \href {https://doi.org/10.18653/v1/D19-1645} {Cross-sentence n-ary
  relation extraction using lower-arity universal schemas}.
\newblock In \emph{Proceedings of the 2019 Conference on Empirical Methods in
  Natural Language Processing and the 9th International Joint Conference on
  Natural Language Processing (EMNLP-IJCNLP)}, pages 6225--6231, Hong Kong,
  China. Association for Computational Linguistics.

\bibitem[{Araki et~al.(2018)Araki, Mulaffer, Pandian, Yamakawa, Oflazer, and
  Mitamura}]{araki2018interoperable}
Jun Araki, Lamana Mulaffer, Arun Pandian, Yukari Yamakawa, Kemal Oflazer, and
  Teruko Mitamura. 2018.
\newblock Interoperable annotation of events and event relations across
  domains.
\newblock In \emph{Proceedings 14th Joint ACL-ISO Workshop on Interoperable
  Semantic Annotation}, pages 10--20.

\bibitem[{Aroyo and Welty(2013)}]{aroyo2013measuring}
Lora Aroyo and Chris Welty. 2013.
\newblock Measuring crowd truth for medical relation extraction.
\newblock In \emph{2013 AAAI Fall Symposium Series}.

\bibitem[{Bartlett et~al.(2006)Bartlett, Fath, Demasi, Hermes, Quinn, Mondou,
  and Rousseau}]{bartlett2006updated}
John~A Bartlett, Michael~J Fath, Ralph Demasi, Ashwaq Hermes, Joseph Quinn,
  Elsa Mondou, and Franck Rousseau. 2006.
\newblock An updated systematic overview of triple combination therapy in
  antiretroviral-naive hiv-infected adults.
\newblock \emph{Aids}, 20(16):2051--2064.

\bibitem[{Beltagy et~al.(2019)Beltagy, Lo, and Cohan}]{scibert}
Iz~Beltagy, Kyle Lo, and Arman Cohan. 2019.
\newblock \href {https://doi.org/10.18653/v1/D19-1371} {{S}ci{BERT}: A
  pretrained language model for scientific text}.
\newblock In \emph{Proceedings of the 2019 Conference on Empirical Methods in
  Natural Language Processing and the 9th International Joint Conference on
  Natural Language Processing (EMNLP-IJCNLP)}, pages 3615--3620, Hong Kong,
  China. Association for Computational Linguistics.

\bibitem[{Bhusal et~al.(2005)Bhusal, Shiohira, and
  Yamane}]{bhusal2005determination}
Y~Bhusal, CM~Shiohira, and N~Yamane. 2005.
\newblock Determination of in vitro synergy when three antimicrobial agents are
  combined against mycobacterium tuberculosis.
\newblock \emph{International journal of antimicrobial agents}, 26(4):292--297.

\bibitem[{Carew et~al.(2008)Carew, Giles, and Nawrocki}]{carew2008histone}
Jennifer~S Carew, Francis~J Giles, and Steffan~T Nawrocki. 2008.
\newblock Histone deacetylase inhibitors: mechanisms of cell death and promise
  in combination cancer therapy.
\newblock \emph{Cancer letters}, 269(1):7--17.

\bibitem[{Cohen(1960)}]{cohen1960coefficient}
Jacob Cohen. 1960.
\newblock A coefficient of agreement for nominal scales.
\newblock \emph{Educational and psychological measurement}, 20(1):37--46.

\bibitem[{DeVita et~al.(1975)DeVita, Young, and Canellos}]{PMID:162854}
VT~DeVita, RC~Young, and GP~Canellos. 1975.
\newblock \href
  {https://doi.org/10.1002/1097-0142(197501)35:1&lt;98::aid-cncr2820350115&gt;3.0.co;2-b}
  {Combination versus single agent chemotherapy: a review of the basis for
  selection of drug treatment of cancer}.
\newblock \emph{Cancer}, 35(1):98—110.

\bibitem[{Eastman and Fidock(2009)}]{eastman2009artemisinin}
Richard~T Eastman and David~A Fidock. 2009.
\newblock Artemisinin-based combination therapies: a vital tool in efforts to
  eliminate malaria.
\newblock \emph{Nature Reviews Microbiology}, 7(12):864--874.

\bibitem[{Fleiss(1971)}]{fleiss1971measuring}
Joseph~L Fleiss. 1971.
\newblock Measuring nominal scale agreement among many raters.
\newblock \emph{Psychological bulletin}, 76(5):378.

\bibitem[{Gu et~al.(2020)Gu, Tinn, Cheng, Lucas, Usuyama, Liu, Naumann, Gao,
  and Poon}]{pubmedbert}
Yu~Gu, Robert Tinn, Hao Cheng, Michael Lucas, Naoto Usuyama, Xiaodong Liu,
  Tristan Naumann, Jianfeng Gao, and Hoifung Poon. 2020.
\newblock \href {http://arxiv.org/abs/arXiv:2007.15779} {Domain-specific
  language model pretraining for biomedical natural language processing}.

\bibitem[{Gururangan et~al.(2020)Gururangan, Marasovi{\'c}, Swayamdipta, Lo,
  Beltagy, Downey, and Smith}]{gururangan-etal-2020-dont}
Suchin Gururangan, Ana Marasovi{\'c}, Swabha Swayamdipta, Kyle Lo, Iz~Beltagy,
  Doug Downey, and Noah~A. Smith. 2020.
\newblock \href {https://doi.org/10.18653/v1/2020.acl-main.740} {Don{'}t stop
  pretraining: Adapt language models to domains and tasks}.
\newblock In \emph{Proceedings of the 58th Annual Meeting of the Association
  for Computational Linguistics}, pages 8342--8360, Online. Association for
  Computational Linguistics.

\bibitem[{Han et~al.(2018)Han, Zhu, Yu, Wang, Yao, Liu, and
  Sun}]{han2018fewrel}
Xu~Han, Hao Zhu, Pengfei Yu, Ziyun Wang, Yuan Yao, Zhiyuan Liu, and Maosong
  Sun. 2018.
\newblock \href {http://arxiv.org/abs/1810.10147} {Fewrel: A large-scale
  supervised few-shot relation classification dataset with state-of-the-art
  evaluation}.

\bibitem[{Hayes and Krippendorff(2007)}]{hayes2007answering}
Andrew~F Hayes and Klaus Krippendorff. 2007.
\newblock Answering the call for a standard reliability measure for coding
  data.
\newblock \emph{Communication methods and measures}, 1(1):77--89.

\bibitem[{Herrero-Zazo et~al.(2013)Herrero-Zazo, Segura-Bedmar, Mart{\'\i}nez,
  and Declerck}]{herrero2013ddi}
Mar{\'\i}a Herrero-Zazo, Isabel Segura-Bedmar, Paloma Mart{\'\i}nez, and
  Thierry Declerck. 2013.
\newblock The {DDI} corpus: An annotated corpus with pharmacological substances
  and drug--drug interactions.
\newblock \emph{Journal of biomedical informatics}, 46(5):914--920.

\bibitem[{Hou et~al.(2019)Hou, Jochim, Gleize, Bonin, and
  Ganguly}]{hou-etal-2019-identification}
Yufang Hou, Charles Jochim, Martin Gleize, Francesca Bonin, and Debasis
  Ganguly. 2019.
\newblock \href {https://doi.org/10.18653/v1/P19-1513} {Identification of
  tasks, datasets, evaluation metrics, and numeric scores for scientific
  leaderboards construction}.
\newblock In \emph{Proceedings of the 57th Annual Meeting of the Association
  for Computational Linguistics}, pages 5203--5213, Florence, Italy.
  Association for Computational Linguistics.

\bibitem[{Ianevski et~al.(2020)Ianevski, Yao, Biza, Zusinaite, Mannik, Kivi,
  Planken, Kurg, Tombak, Ustav et~al.}]{ianevski2020identification}
Aleksandr Ianevski, Rouan Yao, Svetlana Biza, Eva Zusinaite, Andres Mannik,
  Gaily Kivi, Anu Planken, Kristiina Kurg, Eva-Maria Tombak, Mart Ustav, et~al.
  2020.
\newblock Identification and tracking of antiviral drug combinations.
\newblock \emph{Viruses}, 12(10):1178.

\bibitem[{Ianevski et~al.(2021)Ianevski, Yao, Lysvand, Grødeland, Legrand,
  Oksenych, Zusinaite, Tenson, Bjørås, and Kainov}]{v13091768}
Aleksandr Ianevski, Rouan Yao, Hilde Lysvand, Gunnveig Grødeland, Nicolas
  Legrand, Valentyn Oksenych, Eva Zusinaite, Tanel Tenson, Magnar Bjørås, and
  Denis~E. Kainov. 2021.
\newblock \href {https://doi.org/10.3390/v13091768}
  {Nafamostat–interferon-\textalpha combination suppresses sars-cov-2
  infection in vitro and in vivo by cooperatively targeting host tmprss2}.
\newblock \emph{Viruses}, 13(9).

\bibitem[{Jain et~al.(2020)Jain, van Zuylen, Hajishirzi, and
  Beltagy}]{jain2020scirex}
Sarthak Jain, Madeleine van Zuylen, Hannaneh Hajishirzi, and Iz~Beltagy. 2020.
\newblock \href {http://arxiv.org/abs/2005.00512} {Scirex: A challenge dataset
  for document-level information extraction}.

\bibitem[{Kardas et~al.(2020)Kardas, Czapla, Stenetorp, Ruder, Riedel, Taylor,
  and Stojnic}]{kardas-etal-2020-axcell}
Marcin Kardas, Piotr Czapla, Pontus Stenetorp, Sebastian Ruder, Sebastian
  Riedel, Ross Taylor, and Robert Stojnic. 2020.
\newblock \href {https://doi.org/10.18653/v1/2020.emnlp-main.692} {{AxCell}:
  Automatic extraction of results from machine learning papers}.
\newblock In \emph{Proceedings of the 2020 Conference on Empirical Methods in
  Natural Language Processing (EMNLP)}, pages 8580--8594, Online. Association
  for Computational Linguistics.

\bibitem[{Katzir et~al.(2019)Katzir, Cokol, Aldridge, and
  Alon}]{katzir2019prediction}
Itay Katzir, Murat Cokol, Bree~B Aldridge, and Uri Alon. 2019.
\newblock Prediction of ultra-high-order antibiotic combinations based on
  pairwise interactions.
\newblock \emph{PLoS computational biology}, 15(1):e1006774.

\bibitem[{Krallinger et~al.(2017)Krallinger, Rabal, Akhondi, P{\'e}rez,
  Santamar{\'i}a, Rodr{\'i}guez, Tsatsaronis, Intxaurrondo, L{\'o}pez, Nandal,
  van Buel, Chandrasekhar, Rodenburg, L{\ae}greid, Doornenbal, Oyarz{\'a}bal,
  Lourenço, and Valencia}]{Krallinger2017OverviewOT}
Martin Krallinger, Obdulia Rabal, Saber~Ahmad Akhondi, Mart{\'i}n~P{\'e}rez
  P{\'e}rez, Jesus Santamar{\'i}a, Gael~P{\'e}rez Rodr{\'i}guez, Georgios
  Tsatsaronis, Ander Intxaurrondo, Jos{\'e} Antonio~Baso L{\'o}pez, Umesh~K.
  Nandal, Erin~M. van Buel, Anjana Chandrasekhar, Marleen Rodenburg, Astrid
  L{\ae}greid, Marius~A. Doornenbal, Julen Oyarz{\'a}bal, An{\'a}lia Lourenço,
  and Alfonso Valencia. 2017.
\newblock Overview of the {B}io{C}reative {VI} chemical-protein interaction
  track.

\bibitem[{Lee et~al.(2020)Lee, Yoon, Kim, Kim, Kim, So, and Kang}]{biobert}
Jinhyuk Lee, Wonjin Yoon, Sungdong Kim, Donghyeon Kim, Sunkyu Kim, Chan~Ho So,
  and Jaewoo Kang. 2020.
\newblock Bio{BERT}: a pre-trained biomedical language representation model for
  biomedical text mining.
\newblock \emph{Bioinformatics}, 36(4):1234--1240.

\bibitem[{Li et~al.(2016)Li, Sun, Johnson, Sciaky, Wei, Leaman, Davis,
  Mattingly, Wiegers, and lu}]{cdr}
Jiao Li, Yueping Sun, Robin Johnson, Daniela Sciaky, Chih-Hsuan Wei, Robert
  Leaman, Allan~Peter Davis, Carolyn Mattingly, Thomas Wiegers, and Zhiyong lu.
  2016.
\newblock \href {https://doi.org/10.1093/database/baw068} {Biocreative {V}
  {CDR} task corpus: a resource for chemical disease relation extraction}.
\newblock \emph{Database}, 2016:baw068.

\bibitem[{Luan et~al.(2018)Luan, He, Ostendorf, and
  Hajishirzi}]{luan-etal-2018-multi}
Yi~Luan, Luheng He, Mari Ostendorf, and Hannaneh Hajishirzi. 2018.
\newblock \href {https://doi.org/10.18653/v1/D18-1360} {Multi-task
  identification of entities, relations, and coreference for scientific
  knowledge graph construction}.
\newblock In \emph{Proceedings of the 2018 Conference on Empirical Methods in
  Natural Language Processing}, pages 3219--3232, Brussels, Belgium.
  Association for Computational Linguistics.

\bibitem[{Montani and Honnibal(2018)}]{Prodigy:2018}
Ines Montani and Matthew Honnibal. 2018.
\newblock \href {http://arxiv.org/abs/to appear} {Prodigy: A new annotation
  tool for radically efficient machine teaching}.
\newblock \emph{Artificial Intelligence}, to appear.

\bibitem[{Nguyen et~al.(2016)Nguyen, Tannier, Ferret, and
  Besan{\c{c}}on}]{nguyen-etal-2016-dataset}
Kiem-Hieu Nguyen, Xavier Tannier, Olivier Ferret, and Romaric Besan{\c{c}}on.
  2016.
\newblock \href {https://aclanthology.org/L16-1307} {A dataset for open event
  extraction in {E}nglish}.
\newblock In \emph{Proceedings of the Tenth International Conference on
  Language Resources and Evaluation ({LREC}'16)}, pages 1939--1943,
  Portoro{\v{z}}, Slovenia. European Language Resources Association (ELRA).

\bibitem[{Niezni et~al.(2022)Niezni, Taub-Tabib, Harris, Sason-Bauer, Amrusi,
  Azagury, Avrashami, Launer-Wachs, Borchardt, Kusold, Tiktinsky, Hope,
  Goldberg, and Shamay}]{niezni:2022}
Danna Niezni, Hillel Taub-Tabib, Yuval Harris, Hagit Sason-Bauer, Yakir Amrusi,
  Dana Azagury, Maytal Avrashami, Shaked Launer-Wachs, Jon Borchardt, M~Kusold,
  Aryeh Tiktinsky, Tom Hope, Yoav Goldberg, and Yosi Shamay. 2022.
\newblock Extending the boundaries of cancer therapeutic complexity with
  literature data mining.
\newblock \emph{bioRxiv preprint}.

\bibitem[{Peng et~al.(2017)Peng, Poon, Quirk, Toutanova, and tau
  Yih}]{peng2017crosssentence}
Nanyun Peng, Hoifung Poon, Chris Quirk, Kristina Toutanova, and Wen tau Yih.
  2017.
\newblock \href {http://arxiv.org/abs/1708.03743} {Cross-sentence n-ary
  relation extraction with graph lstms}.

\bibitem[{Peng et~al.(2019)Peng, Yan, and Lu}]{bluebert}
Yifan Peng, Shankai Yan, and Zhiyong Lu. 2019.
\newblock Transfer learning in biomedical natural language processing: An
  evaluation of {BERT} and {ELM}o on ten benchmarking datasets.
\newblock In \emph{Proceedings of the 2019 Workshop on Biomedical Natural
  Language Processing (BioNLP 2019)}, pages 58--65.

\bibitem[{Rochlani et~al.(2017)Rochlani, Khan, Banach, and
  Aronow}]{rochlani2017two}
Yogita Rochlani, Mohammed~Hasan Khan, Maciej Banach, and Wilbert~S Aronow.
  2017.
\newblock Are two drugs better than one? a review of combination therapies for
  hypertension.
\newblock \emph{Expert opinion on pharmacotherapy}, 18(4):377--386.

\bibitem[{Rosenman et~al.(2020)Rosenman, Jacovi, and
  Goldberg}]{rosenman2020exposing}
Shachar Rosenman, Alon Jacovi, and Yoav Goldberg. 2020.
\newblock \href {http://arxiv.org/abs/2010.03656} {Exposing shallow heuristics
  of relation extraction models with challenge data}.

\bibitem[{Sabo et~al.(2021)Sabo, Elazar, Goldberg, and
  Dagan}]{sabo2021revisiting}
Ofer Sabo, Yanai Elazar, Yoav Goldberg, and Ido Dagan. 2021.
\newblock \href {http://arxiv.org/abs/2104.08481} {Revisiting few-shot relation
  classification: Evaluation data and classification schemes}.

\bibitem[{Sellam et~al.(2021)Sellam, Yadlowsky, Wei, Saphra, D'Amour, Linzen,
  Bastings, Turc, Eisenstein, Das, Tenney, and Pavlick}]{Sellam2021TheMB}
Thibault Sellam, Steve Yadlowsky, Jason Wei, Naomi Saphra, Alexander D'Amour,
  Tal Linzen, Jasmijn Bastings, Iulia Turc, Jacob Eisenstein, Dipanjan Das, Ian
  Tenney, and Ellie Pavlick. 2021.
\newblock The multiberts: Bert reproductions for robustness analysis.
\newblock \emph{ArXiv}, abs/2106.16163.

\bibitem[{Shlain et~al.(2020)Shlain, Taub-Tabib, Sadde, and
  Goldberg}]{Shlain2020SyntacticSB}
Micah Shlain, Hillel Taub-Tabib, Shoval Sadde, and Yoav Goldberg. 2020.
\newblock Syntactic search by example.
\newblock In \emph{ACL}.

\bibitem[{Shuhendler et~al.(2010)Shuhendler, Cheung, Manias, Connor, Rauth, and
  Wu}]{shuhendler2010novel}
Adam~J Shuhendler, Richard~Y Cheung, Janet Manias, Allegra Connor, Andrew~M
  Rauth, and Xiao~Yu Wu. 2010.
\newblock A novel doxorubicin-mitomycin c co-encapsulated nanoparticle
  formulation exhibits anti-cancer synergy in multidrug resistant human breast
  cancer cells.
\newblock \emph{Breast cancer research and treatment}, 119(2):255--269.

\bibitem[{Taub-Tabib et~al.(2020)Taub-Tabib, Shlain, Sadde, Lahav, Eyal, Cohen,
  and Goldberg}]{taub2020interactive}
Hillel Taub-Tabib, Micah Shlain, Shoval Sadde, Dan Lahav, Matan Eyal, Yaara
  Cohen, and Yoav Goldberg. 2020.
\newblock Interactive extractive search over biomedical corpora.
\newblock \emph{arXiv preprint arXiv:2006.04148}.

\bibitem[{Wasserman et~al.(2001)Wasserman, Sutherland, and
  Cvitkovic}]{wasserman2001irinotecan}
Ernesto Wasserman, William Sutherland, and Esteban Cvitkovic. 2001.
\newblock Irinotecan plus oxaliplatin: a promising combination for advanced
  colorectal cancer.
\newblock \emph{Clinical colorectal cancer}, 1(3):149--153.

\bibitem[{Wu et~al.(2019)Wu, Luo, Leung, Ting, and Lam}]{wu2019renet}
Ye~Wu, Ruibang Luo, Henry~CM Leung, Hing-Fung Ting, and Tak-Wah Lam. 2019.
\newblock Renet: A deep learning approach for extracting gene-disease
  associations from literature.
\newblock In \emph{International Conference on Research in Computational
  Molecular Biology}, pages 272--284. Springer.

\bibitem[{Zhang et~al.(2017)Zhang, Zhong, Chen, Angeli, and
  Manning}]{zhang-etal-2017-position}
Yuhao Zhang, Victor Zhong, Danqi Chen, Gabor Angeli, and Christopher~D.
  Manning. 2017.
\newblock \href {https://doi.org/10.18653/v1/D17-1004} {Position-aware
  attention and supervised data improve slot filling}.
\newblock In \emph{Proceedings of the 2017 Conference on Empirical Methods in
  Natural Language Processing}, pages 35--45, Copenhagen, Denmark. Association
  for Computational Linguistics.

\bibitem[{Zhong and Chen(2021)}]{zhong2021frustratingly}
Zexuan Zhong and Danqi Chen. 2021.
\newblock A frustratingly easy approach for entity and relation extraction.
\newblock In \emph{North American Association for Computational Linguistics
  (NAACL)}.

\end{thebibliography}
\bibliographystyle{acl_natbib}

\newpage
\onecolumn 
\appendix

\section{Appendices}

\subsection{Annotation Guidelines}
\label{guidelines}
\begin{figure*}[h!]
    \centering
    \resizebox{\textwidth}{!}{%
    \includegraphics{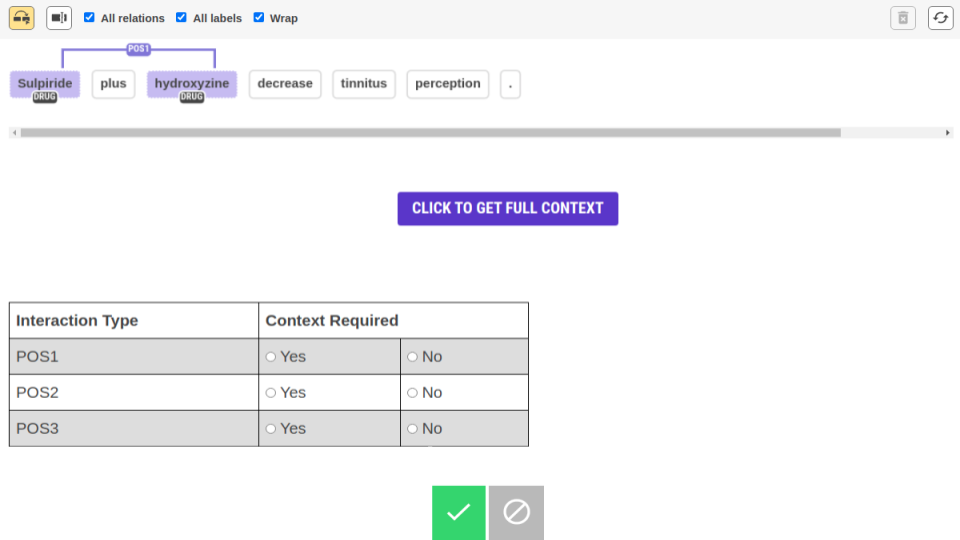}
    }
    \caption{Annotation instance in the Prodigy environment. The screen is constructed of the sentence where they should mark relations, a button to show the full context and a selection per relation to indicate the necessity of the context.}
    \label{fig:annotation}
\end{figure*}

For this task we recruited 7 annotators all studying for advanced degrees in biomedical engineering. The annotators were payed by their advisor, an amount that is standard for annotation projects in their country of residence. All participating annotators were provided with annotation guidelines. The guidelines specified how the annotation process should be carried out and provided definitions and examples for the different labels used. As the task progressed, the guidelines were also expanded to include discussion of frequently encountered issues.

For a given instance, such as presented in the top of \autoref{fig:annotation} the annotator needs to first recognize any missing drugs and mark them, and then label any interactions they find among the drugs. In case they need to consult a wider context they can press on a `show more context' button and a text box with the wider context will appear. This context can be again hidden by clicking the same button if needed. Lastly, in the bottom of the sample page, we present a table with questions regarding the necessity of using the context.

Then the annotator should decide if they need to ignore the current sample or to complete the current instance and accept it, by pressing the accept and ignore buttons.

The annotators are instructed as follows. They should read the sentence carefully, and try to answer a two phase question to themselves. First, if the drugs are mentioned in any form of combination or they should be given separately. Second, if indeed the annotator recognized the drugs as a combination can they determine the efficacy of the combination by the sole sentence.

In case they can not determine the efficacy they are instructed to press on the `get more context' button and read the entire context in order to determine what is the correct efficacy. If after reading the context they can still not determine the efficacy then the label of the interaction should be OTHER\_COMB (aside from negative label experimentation mentioned in Footnote \ref{neg-foot}). Otherwise it should be POS\_COMB. In case that they recognized that there is no combination between the drugs in the sentence then they should not use any label and simply accept the current instance. Then they should answer the context related questions for the POS\_COMB label in order to signal if the context was needed.

While reading the sentence if the annotators find unmarked drugs they can mark them before continuing to the interaction-labeling phase and treat them the same as the other drugs, but, it is not required to mark a word as drug in order to use it in an interaction. If a drug is marked in a wrong manner they should try and fix it, e.g. the span of the drug is incorrect.

In order to achieve more consistent and accurate annotations, they are also instructed to annotate all the interactions that they can find in a given sentence. They should always use the \textit{accept} button even if there are no interactions in the sentence. Only in cases where they want to skip a sentence (e.g. when there is an inherent problem with it) or leave it for a future discussion they should use the \textit{ignore} button. An interaction can occur between more than two drugs, if so they should notice that they don't need each pair from this group to have a marked interaction, as long as they all connect to the same graph. e.g. ``Drugs A, B and C are synergistic.'' connecting A to B and B to C is sufficient, no need to connect drug A to drug C. Each interaction should be marked with a different tag (POS\_COMB1, POS\_COMB2..., OTHER\_COMB1, OTHER\_COMB2...).

\subsection{Evaluation Metric Discussion}
\label{metric}



For measuring the agreement, we chose to use our adaptation of F1 score and not other common metrics such as Cohen's Kappa \citep{cohen1960coefficient} or one of its variations (e.g. Feliss's Kappa \citep{fleiss1971measuring} and Krippendorf's Alpha \citep{hayes2007answering}). These metrics expect a setup where the \textit{relation} candidates are already marked and the task is only to label them -- a labeling task and not an extraction task. This causes two problems, one is that they inherently do not need to handle a partial match. So if for example there are three drugs in a sentence, the first annotator annotated a relation between drugs A and B, while a second annotator annotated the same relation between drugs A, B and C. So we will either underestimate or overestimate their agreement score if we considered this a mismatch or a match respectively. Moreover, their calculations depend on the \textit{hypothetical agreement by chance} normalization factor, but this will not reflect the difficulty of random choosing in our setup as they ignore the size of the combinatorial set of relation candidates we can possibly have.


\subsection{Trigger List}
\label{trigger-list}


\begin{figure*}[!h]
\centering
\resizebox{\textwidth}{!}{
    \begin{tikzpicture}
        \begin{axis}[
            ybar,
            ymin=0,
            enlarge x limits=0.08,
            xlabel={Trigger words},
            ylabel={Percentage of Containing Abstracts},
            symbolic x coords={combination, plus, combined, followed by, first-line, combinations, prior to, synergistic, beneficial, combining, sequential, additive, synergy, first line, synergism},
            xtick=data,
            nodes near coords,
            every node near coord/.append style={font=\small},
            nodes near coords align={vertical},
            x tick label style={rotate=45, anchor=north east, inner sep=0mm, font=\small},
            legend cell align={left},
            xlabel near ticks,
            ylabel near ticks,
            ytick align=outside,
            ytick pos=left,
            yticklabel={\pgfmathprintnumber\tick\%},
            xtick align=inside,
            xtick style={draw=none},
            ylabel shift=-3 pt,
            height=7.5cm,
            width=12.2cm,
            xtick distance=1,
            ]
            \addplot[colorone!20!black,fill=colorone!80!white] coordinates {
    (combination, 49.9)
    (plus, 23)
    (combined, 19.6)
    (followed by, 16.7)
    (first-line, 11.6)
    (combinations, 6.5)
    (prior to, 5.5)
    (synergistic, 4.8)
    (beneficial, 4)
    (combining, 3.9)
    (sequential, 3.5)
    (additive, 2.6)
    (synergy, 1.8)
    (first line, 1.5)
    (synergism, 1)
};
        \end{axis}
        
    \end{tikzpicture}
    }
    \caption{Abstracts percentage including each trigger word (1634 abstracts included; 44 words in the full word list; Words <1\% were neglected from the figure.}
    \label{fig:triggers}
\end{figure*}
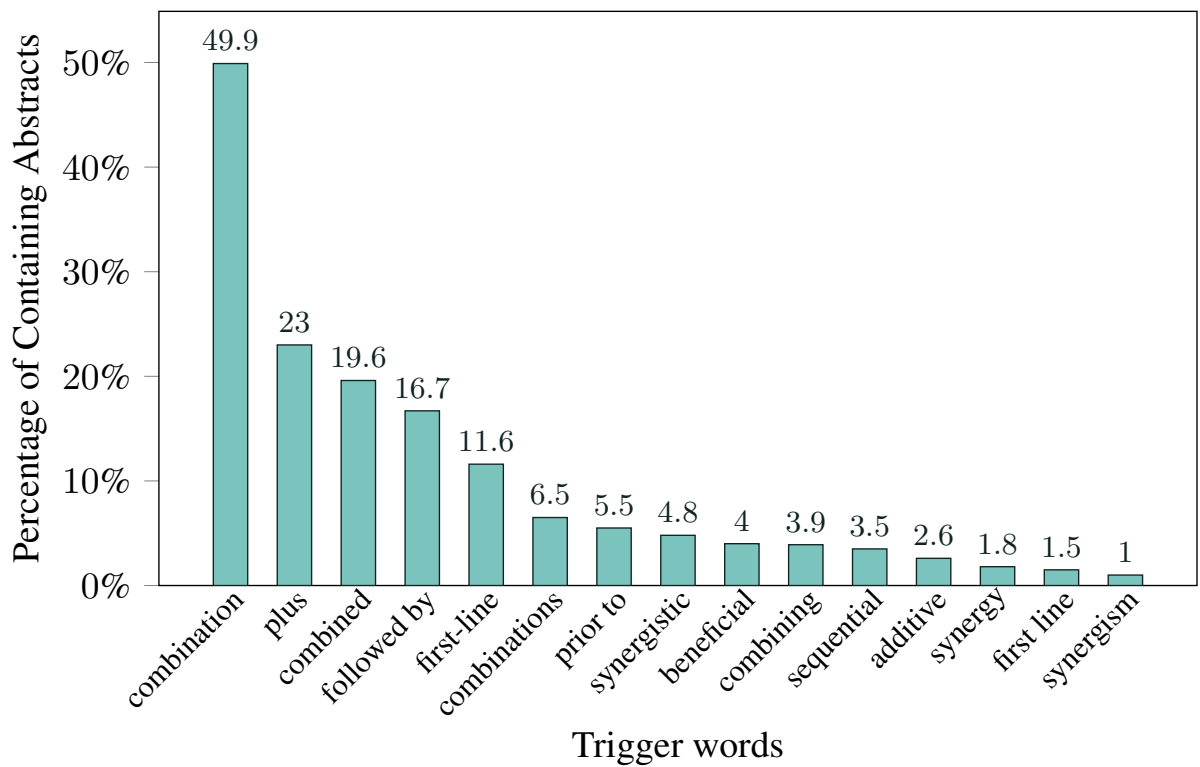

In \autoref{fig:triggers} we show the triggers that we used in the Spike queries. We show the percentage of abstracts that included each trigger (others under 1\%: \textit{conjunction, two-drug, first choice, additivity, combinational, synergetic, simultaneously with, supra-additive, five-drug, combinatory, over-additive, timed-sequential, co-blister, super-additive, synergisms, synergic, synergistical, less-than-additive, greater-than-additive, additivesynergistic, supraadditive, superadditive, overadditive, subadditive, first-choice, 2-drug, sub-additive, more-than-additive, 3-drug}).

\end{document}